# A Survey of Recent Abstract Summarization Techniques*

Diyah Puspitaningrum[1][0000-0001-5442-144X]

[1] University of Bengkulu, Bengkulu 38371A, Indonesia
`diyahpuspitaningrum@gmail.com`

**Abstract.** This paper surveys several recent abstract summarization methods: T5, Pegasus, and ProphetNet. We implement the systems in two languages: English and Indonesian languages. We investigate the impact of pre-training models (one T5, three Pegasuses, three ProphetNets) on several Wikipedia datasets in English and Indonesian language and compare the results to the Wikipedia systems' summaries. The T5-Large, the Pegasus-XSum, and the ProphetNet-CNNDM provide the best summarization. The most significant factors that influence ROUGE performance are coverage, density, and compression. The higher the scores, the better the summary. Other factors that influence the ROUGE scores are the pre-training goal, the dataset's characteristics, the dataset used for testing the pre-trained model, and the cross-lingual function. Several suggestions to improve this paper's limitation are: 1) assure that the dataset used for the pre-training model must sufficiently large, contains adequate instances for handling cross-lingual purpose; 2) Advanced process (finetuning) shall be reasonable. We recommend using the large dataset consists of comprehensive coverage of topics from many languages before implementing advanced processes such as the train-infer-train procedure to the zero-shot translation in the training stage of the pre-training model.

## 1     Introduction

### 1.1     The Goal of The Research

The abstract summarization based on the sequence-to-sequence technique of encoder-decoder (or, Transformers) has recently become a widely known technique. This paper comprises a survey, exploration, and empirical comparison on pre-trained models on English and Indonesian datasets consisting of hundreds and thousands of sentences, respectively, scraped from the web. So far, there is not exist yet any research about Indonesian language summaries based on Transformers. In this work, for evaluation, we dub the coverage, density, and compress metrics [4] to gain insights into the summarization results using ROUGE metrics as in [10].

---

## 2    Preliminaries

### 2.1    Data

The Wikipedia datasets were collected using search engine metadata. We performed a Web-scale crawling of over five topics of Wikipedia pages of each language (English, Indonesian): internet, geography, archaeology, artificial intelligence, and ecology. See Table 1. From each Wikipedia page, from a given query, we provide a gold standard summary of the page by extracting the available summary shown by default of the Wikipedia-API. The Wikipedia metadata we are concerned about includes URL, Title, Description/Body Text, and Summary.

**Table 1.** Statistics of Wikipedia datasets.

| Dataset | Language | Number of Wikipedia Page | Number of Sentences | Number of Tokens |
|---|---|---|---|---|
| Internet | EN | 18 | 9500 | 133047 |
| Internet | ID | 18 | 1637 | 21038 |
| Geography | EN | 423 | 85819 | 1578031 |
| Geography | ID | 423 | 85819 | 1578031 |
| Archaeology | EN | 46 | 9692 | 180752 |
| Archaeology | ID | 46 | 4437 | 53319 |
| Artificial Intelligence | EN | 241 | 809 | 19104 |
| Artificial Intelligence | ID | 241 | 18243 | 462480 |
| Ecology | EN | 291 | 109144 | 1337758 |
| Ecology | ID | 291 | 29226 | 491946 |

EN: English,    ID: Indonesian Language

### 2.2    Definitions

We lack systematic evaluation for abstractive summarizations across diverse domains. In this paper, we borrow extractive fragment coverage and fragment density [4].

**Fragment Coverage.** The fragment coverage formulated as:

$$\text{COVERAGE}(A, S) = \frac{1}{|S|} \sum_{f \in F(A,S)} |f| \qquad (1)$$

Where A is an article (the Wikipedia page), and S is a summary. The fragment coverage measures the percentage of words in a summary derived from a source text (the body text of a Wikipedia page).

**Fragment Density.** The fragment density formulated as:

$$\text{DENSITY}(A, S) = \frac{1}{|S|} \sum_{f \in F(A,S)} |f|^2 \qquad (2)$$

Given a series of extractions, the fragment density measures how well the word sequence. If a summary contains many individual words, it is good because the individual words are adaptive in the new arrangement of sentences. The latest is useful for abstract summarization. It is the average length of the fragment to which each word in the summary belongs. Many individual words from the Wikipedia body text can also arrange a summary composition. The latest conditions cause a high coverage score. However, the situations also open a new arrangement of sentences in summary with new words that previously are not available in the Wikipedia page's body text.

**Compression.** We also dub an idea of compression. Higher compression in any summary means the summary captures many essential sights of the Wikipedia body text. The compression ratio defined as:

$$\text{COMPRESSION}(A, S) = \frac{|A|}{|S|} \tag{3}$$

### 2.3 Basic Concepts

Instead of using RNN for transfer learning on many NLP tasks, the Transformer models are recently widely known. See [17] for the original work of the Transformer. Furthermore, such Sequence-to-Sequence pre-trained models based on Transformer architecture showed success in the text generation, including summarizing (abstract summarizing task) ([2][16][15][8][13]). The original Transformer has an encoder-decoder architecture (see the architecture on [17] Fig.1). The model does sequence-to-sequence tasks using a weighted sum average of the sequence called the "self-attention" [1]. We will explore its implementation on T5, Pegasus, and ProphetNet.

#### 2.3.1 T5 (Text-to-Text Transfer Transformer)

The basic idea of T5 is the "text-to-text" problem. The T5 text-to-text framework applies the same model (a unified model), objective, training procedure, and decoding process to every task of a wide variety of English-based NLP problems (classification, summarization, translation, and question answering). The unified approach's flexibility also benefited from seizing up a model's size to billion parameters and pre-trained corpus. The T5 system was pre-trained using arbitrarily text, which corrupted, and it spans on mask ratios and the dimensions of the sizes of the text spans. See [13] Fig.1 for the diagram of the text-to-text framework.

From the T5 original work [13], they experiment with different approaches: the Language Modeling (predict the afterword in a sentence by considering all preceding words), Deshuffling (shuffle all the words in a sentence followed by training the model to predict the original text), and Denoising objectives (masking a sequence of words from the sentence followed by training the model to predict the masked words). From the three, the denoising objective shows the most promising results.

In T5, its pre-training objective is a "fill-in-the-blank" task (the model predicts missing words within an incomplete piece of text). To do that, T5 created a brand-new downstream task, the sized fill-in-the-blank, where the model must substitute a blank to a specified number of words. T5 will train the input text to fill in the blank with N-words approximately, and then finetuning results using the large dataset to produce realistic outputs. In the T5 original work [13], the "Colossal Clean Crawled Corpus" (C4) used contains hundreds of gigabytes of English text.

### 2.3.2 Theory of Pegasus

It was inspired by T5 work, which success with its masking words and contiguous spans [13], in Pegasus, a pseudo-summary created by picking and masking whole sentences from documents and then concatenating the gap-sentences. Pegasus choose sentences that are important to the documents. Pegasus removed/masked the critical sentences from an input document, and then the critical sentences are generated simultaneously as one output sequence from the other unselected sentences. Pegasus architecture is a kind of Transformer encoder-decoder. The pre-training in Pegasus uses extracted gap-sentences, sequence-to-sequence, to do abstractive summarizing. The gap sentences ratio here is the same as the mask rate of a text document. See Pegasus architecture in [19].

### 2.3.3 Theory of ProphetNet

ProphetNet [12] is also a Sequence-to-sequence pre-training model based on the encoder-decoder Transformer of [17]. The characteristics of ProphetNet are as follow: (1) There is a prophecy characteristic called the future n-gram prediction. On the decoder step, the ProphetNet simultaneously predicts n future tokens by resulting n probability at every time of the steps. We can adjust the weights attenuation between the traditional language modeling and *the future n-gram prediction*. (2) There is an *n*-stream self-attention mechanism. The mechanism is the same as the masked multi-head self-attention in the Transformer decoder. To improve positional information of the decoder, the ProphetNet combines the absolute positional embedding and T5. In ProphetNet, the *i*-th predicting stream predicts the future stream, $y_t$, based on the previous tokens, viz. $y_{<t-i+1}$. The predicting and the main decoder stream shared during training. In predicting, the stream initialized with special tokens. The ProphetNet acts the same as a traditional Transformer decoder if the predicting stream feature is disabled. (3) The third difference is the availability of the activity masking based autoencoder denoising task for Sequential-to-sequential pre-training. In ProphetNet, we mask out some token spans of the original text as the encoder input. ProphetNet recovers the next n future tokens within each masked token span.

## 3 The Experiments

We first describe the T5-Large pre-trained model's details in Section 3.1, followed by the Pegasus-XSum pre-trained model in Section 3.2 ProphetNet-XGlue-NTG in Section 3.3, Pegasus-Multi_news in Section 3.4, Pegasus-Wikihow in Section 3.5, ProphetNet-CNNDM in Section 3.6, and ProphetNet-Squad-QG in Section 3.7.

### 3.1 T5-Large (T5)

To produce T5 summaries, T5 truncates all datasets until 512 words. Other preprocessing is removing newlines. Here, for tokenizer as well as the pre-trained model, we use T5-Large. After being tokenized/encoded, we feed the text to the T5 model with the number of beams=4, and n-gram size=2 can only occur once, the minimum length of 30, the maximum size of 100, and early stopping are True. The T5 decoded back the text to produce output.

The T5-Large used the C4 (Colossal Clean Crawled Corpus) dataset to build a pre-trained model. It is a pre-train language model on massive unlabeled datasets. For C4, taken from Common Crawl scrape from April 2019 and applied some cleansing filters, it results in a very clean 750GB text dataset of large pre-training datasets, more extensive than other pre-training datasets.

### 3.2 Pegasus-XSum (Pegasus)

To produce the Pegasus-XSum summaries, the Pegasus truncates all datasets until 512 words. Other preprocessing is removing newlines. Here, for the tokenizer and the pre-trained model, we use Pegasus-XSum. The procedures are as follows: After being tokenized/encoded, we feed the text to the Pegasus model by token truncation and pad to the batch's most extended/longest sequence as output. We then decode the tokenized output by skipping special tokens as the target tokens. The latest act as an output. The Pegasus-XSum dataset used the 227k BBC articles from 2010 to 2017 from various subjects to build a pre-trained model. See [11].

### 3.3 ProphetNet-XGlue-NTG (ProphetNet)

NTG is short for News Titles Generation. To produce the ProphetNet summaries, the ProphetNet truncated all datasets until 512 words. Other preprocessing is removing newlines. Here, for the tokenizer and the pre-trained model, we use the ProphetNet-XGlue-NTG. We tokenized with a maximum length of padding the sentences = 256. After being tokenized/encoded, we feed the text to the ProphetNet model with the number of beams=4, the maximum length of the sequence to be generated = 100, and early stopping is True. We then convert the output of lists of token ids into a list of strings by removing special tokens in the decoding. We then feed the output again to the tokenizer, with a maximum length of padding the sentences = 256. The result then becomes an input in the model, and then the output generated from the model is fed into the tokenizer again by skipping special tokens. The latest contains the target text or the (decoded) output.

The ProphetNet-XGlue-NTG is a cross-lingual version ProphetNet, pre-trained on wiki100 xGLUE dataset and finetuned on XGLUE cross-lingual News Titles Generation task. ProphetNet can predict more future tokens with an n-stream decoder. For XGLUE cross-lingual NTG tasks, the Transformers finetuned the ProphetNet with English data, but inference with both English and other zero-shot language data. Comparing to GLUE [18], which label in English for natural language understanding tasks only, XGLUE [9], has two main advantages. First, it unified 11 NLP tasks consist of: named entity recognition, POS tagging, text classification, multilingual question answering, cross-lingual natural language inference corpus, a cross-lingual adversarial dataset for paraphrase identification, query-ad matching, web page ranking, question answering matching, question generation, and news title generation. Second, for every task, it gives labeled data in many different languages.

### 3.4 Pegasus-Multi_News (Pegasus)

To produce the Pegasus-Multi_News summaries, the Pegasus truncated all datasets until 512 words. Other preprocessing is removing newlines. Here, for the tokenizer and the pre-trained model, we use Pegasus-Multi_News. We tokenized with a maximum length of padding the sentences = 256. After being tokenized/encoded, we feed the text to the Pegasus model with the number of beams=4, the maximum length of the sequence generated by the Pegasus = 100, and early stopping is True. We then convert the output of lists of token ids into a list of strings by removing special tokens in the decoding. We then feed the result again to the tokenizer, with a maximum length of padding the sentences = 256. The output then becomes an input in the model, and then the output generated from the model is fed into the tokenizer again by skipping special tokens. The latest contains the target text, or we called the (decoded) output. The Multi-News dataset [3] is a multiple document summarization dataset consists of 56k (news, human-written summaries) pairs.

### 3.5 Pegasus-Wikihow (Pegasus)

To produce the Pegasus-Wikihow summaries, the Pegasus truncated all datasets until 512 words. Other preprocessing is removing newlines. For the tokenizer and the pre-trained model, we use the Wikihow dataset for Pegasus-Wikihow. We tokenized with a maximum length of padding the sentences = 256. After being tokenized/encoded, we feed the text to the Pegasus model with the number of beams=4, the maximum length of the sequence generated by Pegasus = 100, and early stopping is True. We convert the output of lists of token ids into a list of strings by removing special tokens in the decoding. We then feed the result again to the tokenizer, with a maximum length of padding the sentences = 256. The result then becomes an input in the model. The output generated from the model is fed into the tokenizer again by skipping special tokens. The latest contains the target text, or we called the (decoded) output. The Wikihow dataset [6] is the concatenated summary-sentences from multiple instruction-step paragraphs. Each of the 200k instances consisting of instruction paragraphs and its summary.

### 3.6 ProphetNet-CNNDM (ProphetNet)

CNNDM is short for CNN/Daily Mail newspaper articles. To produce the Prophet Net-CNNDM summaries, the ProphetNet truncated all datasets until 512 words. Other preprocessing is removing newlines. Here, we use the ProphetNet-CNNDM for the tokenizer and the pre-trained model. We tokenized with a maximum length of padding the sentences = 100. After being tokenized/encoded, we feed the text to the ProphetNet model with the number of beams=4, the maximum length of the sequence to be generated = 512, and early stopping is True. We then convert the output of lists of token ids into a list of strings by removing special tokens in the decoding. The latest contains the target text, or we called the (decoded) output. CNNDM dataset consists of 93k CNN articles and 220k Daily Mail articles [5].

### 3.7 ProphetNet-Squad-QG (ProphetNet)

QG is short for Question Generation. To produce the ProphetNet-Squad-QG summaries, the ProphetNet truncated all datasets until 512 words. Other preprocessing is removing newlines. Here, for tokenizer and the pre-trained model, we use ProphetNet-Squad-QG. We tokenized with a maximum length of padding the sentences = 100. After being tokenized/encoded, we feed the text to the ProphetNet model with the number of beams=4, the maximum length of the sequence to be generated = 512, and early stopping is True. We then convert the output of lists of token ids into a list of strings by removing special tokens in the decoding. The latest contains the target text, or we called the (decoded) output. SQUAD (Stanford Question Answering Dataset) is a reading comprehension dataset consists of questions on Wikipedia articles set and their answers [14]. QG is a data-driven question generation model [18].

## 4 Results

We survey on one T5 model (T5-Large), three variants of Pegasus (Pegasus-XSum, Pegasus-Multi_News, Pegasus-Wikihow), and three variants of ProphetNet (ProphetNet-XGlue-NTG, ProphetNet-CNNDM, ProphetNet-Squad-QG). Since our goal is to compare various pre-trained models to find which pre-trained model works best with English and Indonesian languages, we report the models' performance on the ROUGE metrics for all summary tasks as in [10]. We also interest in analyzing the execution time as well as the best summary results. For all tables in this paper, T5 means T5-Large, Peg_xsum means Pegasus-XSum, PN-xglue-ntg is ProphetNet-XGlue-NTG, Peg_multi_news is Pegasus-Multi_News, Peg_wikihow means Pegasus-Wikihow, PN_cnndm is ProphetNet-CNNDM, and PN_squad_qg is ProphetNet-Squad-QG.

**The Fragmental metric.** Table 2 shows the fragmental metric of coverage, density, and compression on English and Indonesian languages. We obtain the scores by computing the coverage, density, and compression of a summary and a Wikipedia page's body text. The coverage scores are the total coverage divides by the number of records in a dataset (the total number of Wikipedia pages, in our case). Further, the density scores are the total density divides by the number of records in a dataset. The compression scores compute the total compression divides by the number of records in a dataset. Table 2 shows that compared to Wikipedia's body text page, the 'Wiki' (the Wikipedia summary given by default of Wikipedia-API) has excellent fragmental coverages. Some even reach above 99% coverages. The density scores are also high. The 'Wiki' in here plays the role of the gold standard.

**Table 2.** The fragmental metric of coverage, density, and compression. (a) English dataset (left, code: EN), (b) Indonesian language dataset (right, code: ID).

| Dataset | Method | Language | Coverage | Density | Compress | Language | Coverage | Density | Compress |
|---|---|---|---|---|---|---|---|---|---|
| Internet | Wiki | EN | 93.81 | 4521.6 | 4616.73 | ID | 79.38 | 4400.52 | 1185.45 |
| Geography | Wiki | EN | 94.88 | 7073.52 | 2440.87 | ID | 95.18 | 8255.36 | 1095.77 |
| Archaeology | Wiki | EN | 92.15 | 5581.07 | 1986.07 | ID | 98.95 | 12944.87 | 497.05 |
| AI | Wiki | EN | 99.71 | 11403.94 | 2117.79 | ID | 98.73 | 9138.66 | 883.81 |
| Ecology | Wiki | EN | 99.8 | 13629.79 | 1644.4 | ID | 99.61 | 9873.54 | 625.37 |
| Internet | T5 | EN | 99.31 | 1270.69 | 15720.17 | ID | 97.81 | 765.47 | 5406.79 |
| Geography | T5 | EN | 97.45 | 1176.67 | 7173.86 | ID | 96.48 | 738.18 | 4692.69 |
| Archaeology | T5 | EN | 98.36 | 1498.98 | 8363.67 | ID | 96.57 | 679.71 | 6486.05 |
| AI | T5 | EN | 97.84 | 1393.92 | 6663.45 | ID | 96.95 | 788.01 | 5960.89 |
| Ecology | T5 | EN | 97.87 | 1409.36 | 5978.81 | ID | 97 | 745.36 | 5310.03 |
| Internet | Peg_xsum | EN | 81.25 | 367.5 | 64698.86 | ID | 80.67 | 492.96 | 24382.54 |
| Geography | Peg_xsum | EN | 78.77 | 667.95 | 36645.65 | ID | 79.16 | 415.45 | 6127.47 |
| Archaeology | Peg_xsum | EN | 67.9 | 180.93 | 42948.91 | ID | 91.39 | 514.91 | 7344.01 |
| AI | Peg_xsum | EN | 88.6 | 1050.45 | 19659.18 | ID | 89.6 | 497.21 | 8111.5 |
| Ecology | Peg_xsum | EN | 89.46 | 1081.1 | 21669.94 | ID | 90.23 | 560.06 | 7023.78 |
| Internet | PN-xglue-ntg | EN | 88.13 | 290.9 | 102985.3 | ID | 66.94 | 142.82 | 24450.51 |
| Geography | PN-xglue-ntg | EN | 79.46 | 266.22 | 50550.49 | ID | 65.57 | 220.29 | 18796.08 |
| Archaeology | PN-xglue-ntg | EN | 82.39 | 218.46 | 56684.76 | ID | 79.23 | 165.85 | 26150.88 |
| AI | PN-xglue-ntg | EN | 91.41 | 260.8 | 42027.21 | ID | 69.84 | 156.66 | 19950.59 |
| Ecology | PN-xglue-ntg | EN | 89.3 | 250.27 | 39069.01 | ID | 71.2 | 189.4 | 19856.08 |
| Internet | Peg_multi_news | EN | 71.5 | 141.22 | 10492.47 | ID | 36.88 | 42.96 | 2013.72 |
| Archaeology | Peg_multi_news | EN | 62.46 | 143.34 | 5598.09 | ID | 36.97 | 44.92 | 1925.02 |
| Internet | Peg_wikihow | EN | 81.4 | 183.23 | 26810.68 | ID | 56.51 | 145.24 | 9422.9 |
| Archaeology | Peg_wikihow | EN | 71.05 | 132.51 | 15135.65 | ID | 51.15 | 116.64 | 8274.44 |
| Internet | PN_cnndm | EN | 91.76 | 947.76 | 26075.21 | ID | 65.74 | 443.56 | 5343.27 |
| Archaeology | PN_cnndm | EN | 88.95 | 999.88 | 10336.65 | ID | 70.48 | 551.08 | 5283.4 |
| Internet | PN_squad_qg | EN | 79.02 | 162.87 | 82716.96 | ID | 40.81 | 59.4 | 14509.65 |
| Archaeology | PN_squad_qg | EN | 73.05 | 181.51 | 35673.88 | ID | 39.41 | 70.94 | 14921.39 |
| AI: Artificial Intelligence | | Max | 99.8 | 13629.79 | 102985.3 | Max | 99.61 | 12944.87 | 26150.88 |

T5: T5-Large, Peg_xsum: Pegasus-XSum, PN-xglue-ntg: ProphetNet-XGlue-NTG, Peg_multi_news: Pegasus-Multi_News, Peg_wikihow: Pegasus-Wikihow, PN_cnndm: ProphetNet-CNNDM, PN_squad_qg: ProphetNet-Squad-QG

**The ROUGE Scores.** ROUGE-1-F computes the overlap of unigrams (one single word) between the system summary (candidate summary) and the reference summaries. On the other hand, ROUGE-2-F computes bigrams' overlap (a bigram is two consecutive words) between the system summary and the reference summaries. ROUGE-L-F computes the longest matching sequence of words using LCS (Longest Common Subsequence). It is unnecessary to have successive matches but matches in sequence in the sentence that reflects word order. Two compared summaries are more similar if they have a long common sequence of words than any other.

Recall (*r*) refers to the number of overlapping words divided by total words in a reference summary. Reference summary here means the gold standard summary. Precision (*p*) refers to the number of overlapping words divided by total words in the candidate summary. Candidate summary here means any other summaries other than the gold standard. F-measure (*f*) provides complete information involving the recall and the precision as follows: $f = \frac{(1+\beta^2)r \cdot p}{r+\beta^2 \cdot p}$. The β value chosen such that recall considered β times as important as precision. When β=0, the *f* favors *r*; when β=1, the *f* favors *p*. In this paper, we use the ROUGE default, β=0.5.

**Table 3.** Rouge performance for the Wikipedia English dataset.

| Dataset | Lang | #rows | Method | rouge-1 | | | rouge-2 | | | rouge-l | | | Final |
|---|---|---|---|---|---|---|---|---|---|---|---|---|---|
| | | | | f | p | r | f | p | r | f | p | r | Metric |
| internet | en | 18 | T5 | 29.47 | 71.16 | 20.53 | 20.25 | 51.3 | 13.65 | 36.09 | 67.95 | 26.66 | 28.6 |
| geography | en | 423 | T5 | 29.91 | 67.04 | 21.61 | 19.72 | 45.97 | 14.01 | 35.64 | 62.28 | 27.3 | 28.42 |
| archaeology | en | 46 | T5 | 27.65 | 73.39 | 18.69 | 18.59 | 51.56 | 12.03 | 32.26 | 67.99 | 22.53 | 26.17 |
| artificial intelligence | en | 241 | T5 | 34.41 | 77.91 | 25.21 | 26.09 | 60.91 | 18.95 | 42.74 | 74.83 | 32.97 | 34.41 |
| ecology | en | 291 | T5 | 29.92 | 76.26 | 20.63 | 21.92 | 58.51 | 14.85 | 37.12 | 72.24 | 27.14 | 29.65 |
| internet | en | 18 | Peg_xsum | 10.28 | 55.06 | 6.33 | 4.59 | 19.73 | 2.87 | 12.32 | 51.81 | 7.89 | 9.06 |
| geography | en | 423 | Peg_xsum | 11.49 | 58.91 | 7.08 | 6.38 | 28.53 | 4 | 13.98 | 54.79 | 8.81 | 10.62 |
| archaeology | en | 46 | Peg_xsum | 6.09 | 54.76 | 3.58 | 2.09 | 18.36 | 1.26 | 7.74 | 51.26 | 4.55 | 5.31 |
| artificial intelligence | en | 241 | Peg_xsum | 15.45 | 67.65 | 9.64 | 10.39 | 45.43 | 6.47 | 19.69 | 63.19 | 12.66 | 15.18 |
| ecology | en | 291 | Peg_xsum | 13.65 | 66.78 | 8.38 | 9.42 | 43.18 | 5.87 | 17.59 | 62.85 | 11.13 | 13.55 |
| internet | en | 18 | PN-xglue-ntg | 6.47 | 64.02 | 3.58 | 2.84 | 34.61 | 1.53 | 8.31 | 59.69 | 4.68 | 13.55 |
| geography | en | 423 | PN-xglue-ntg | 5.61 | 57.61 | 3.08 | 2.13 | 22.9 | 1.17 | 7.88 | 55.01 | 4.4 | 5.21 |
| archaeology | en | 46 | PN-xglue-ntg | 4.48 | 58.09 | 2.37 | 1.34 | 20.1 | 0.7 | 5.97 | 54.81 | 3.19 | 3.93 |
| artificial intelligence | en | 241 | PN-xglue-ntg | 8.02 | 75.36 | 4.47 | 4.16 | 44.13 | 2.29 | 11.4 | 73.82 | 6.47 | 7.86 |
| ecology | en | 291 | PN-xglue-ntg | 6.27 | 74.68 | 3.4 | 3.21 | 40.62 | 1.74 | 9.03 | 71.81 | 4.96 | 6.17 |
| internet | en | 18 | Peg_multi_news | 14 | 26.47 | 11.31 | 1.39 | 3.01 | 1.03 | 11.55 | 19.85 | 9.06 | 8.98 |
| archaeology | en | 46 | Peg_multi_news | 17.58 | 33.23 | 14.29 | 2.14 | 5.25 | 1.57 | 12.83 | 21.31 | 10.48 | 10.85 |
| internet | en | 18 | Peg_wikihow | 11.69 | 43.1 | 7.35 | 2.51 | 8.91 | 1.57 | 12.59 | 37.33 | 8.16 | 8.93 |
| archaeology | en | 46 | Peg_wikihow | 11.89 | 45.69 | 7.55 | 2.1 | 9.47 | 1.3 | 11.63 | 38.15 | 7.35 | 8.54 |
| internet | en | 18 | PN_cnndm | 24.16 | 66.83 | 16.79 | 18.51 | 46.33 | 13.2 | 28.33 | 65.29 | 20.5 | 23.67 |
| archaeology | en | 46 | PN_cnndm | 22.75 | 71.7 | 14.3 | 17.64 | 55.12 | 11.07 | 28.42 | 67.63 | 18.95 | 22.94 |
| internet | en | 18 | PN_squad_qg | 3.27 | 41.25 | 1.74 | 1.05 | 15.17 | 0.55 | 5.02 | 40.66 | 2.76 | 3.11 |
| archaeology | en | 46 | PN_squad_qg | 3.6 | 39.32 | 1.97 | 1.13 | 14.79 | 0.61 | 5.11 | 38.07 | 2.85 | 3.28 |
| | | | MAX | 34.41 | 77.91 | 25.21 | 26.09 | 60.91 | 18.95 | 42.74 | 74.83 | 32.97 | 34.41 |

Table 3 shows ROUGE performance for the Wikipedia English dataset. While Table 4 for the Indonesian language dataset. From Table 3 and Table 4, we can see that in general, *p* is more significant than *r*. If the ROUGE score for *p* has a lower value, it means that some of either the unigram, or the bigram, or the LCS word(s), were found repetitive in the candidate summary. If such repetitions occur in the reference summary, then it will lower the recall scores.

Table 4. Rouge performance for the Wikipedia Indonesian language dataset.

| Dataset | Lang | #rows | Method | rouge-1 | | | rouge-2 | | | rouge-l | | | Final |
|---|---|---|---|---|---|---|---|---|---|---|---|---|---|
| | | | | f | p | r | f | p | r | f | p | r | Metric |
| internet | id | 18 | T5 | 15.94 | 53.69 | 10.36 | 9.27 | 35.88 | 5.58 | 20.77 | 52.55 | 14.03 | 15.33 |
| geography | id | 423 | T5 | 20.41 | 61.85 | 13.54 | 12.95 | 41.31 | 8.46 | 26.85 | 59.32 | 18.9 | 20.07 |
| archaeology | id | 46 | T5 | 14.57 | 81.07 | 8.34 | 9.69 | 58.38 | 5.49 | 21.59 | 79 | 13.03 | 15.29 |
| artificial intelligence | id | 241 | T5 | 19.95 | 79.76 | 12.24 | 13.99 | 58.97 | 8.54 | 27.96 | 77.16 | 18.19 | 20.63 |
| ecology | id | 291 | T5 | 19.67 | 80.51 | 12.16 | 13.95 | 60.1 | 8.61 | 27.9 | 78.18 | 18.18 | 20.51 |

| Dataset | Lang | #rows | Method | rouge-1 | | | rouge-2 | | | rouge-l | | | Final |
|---|---|---|---|---|---|---|---|---|---|---|---|---|---|
| | | | | f | p | r | f | p | r | f | p | r | Metric |
| internet | id | 18 | Peg_xsum | 15.99 | 55.19 | 10.13 | 7.34 | 25.29 | 4.58 | 19.3 | 54.82 | 12.58 | 14.21 |
| geography | id | 423 | Peg_xsum | 14.02 | 39.52 | 9.59 | 4.52 | 12.91 | 3.06 | 14.19 | 38.21 | 9.53 | 10.91 |
| archaeology | id | 46 | Peg_xsum | 17.79 | 83.53 | 10.45 | 11.86 | 58.16 | 6.94 | 21.41 | 85.57 | 12.63 | 17.02 |
| artificial intelligence | id | 241 | Peg_xsum | 20.42 | 82.75 | 12.59 | 13.98 | 57.73 | 8.63 | 25.93 | 83.69 | 16.37 | 20.11 |
| ecology | id | 291 | Peg_xsum | 19.99 | 85.32 | 12.2 | 13.74 | 59.38 | 8.38 | 25.72 | 86.75 | 16.11 | 19.82 |

| Dataset | Lang | #rows | Method | rouge-1 | | | rouge-2 | | | rouge-l | | | Final |
|---|---|---|---|---|---|---|---|---|---|---|---|---|---|
| | | | | f | p | r | f | p | r | f | p | r | Metric |
| internet | id | 18 | PN-xglue-ntg | 6.68 | 45.47 | 3.86 | 3.22 | 17.95 | 1.9 | 8.28 | 44.53 | 4.86 | 6.06 |
| geography | id | 423 | PN-xglue-ntg | 3.93 | 32.99 | 2.19 | 1.15 | 9.65 | 0.64 | 5.38 | 31.88 | 3.06 | 3.49 |
| archaeology | id | 46 | PN-xglue-ntg | 4.29 | 66.69 | 2.24 | 2.09 | 32.65 | 1.09 | 6.11 | 61.83 | 3.24 | 4.16 |
| artificial intelligence | id | 241 | PN-xglue-ntg | 5.82 | 55.6 | 3.18 | 3.03 | 27.42 | 1.66 | 8.23 | 54.49 | 4.6 | 5.69 |
| ecology | id | 291 | PN-xglue-ntg | 6.69 | 58 | 3.72 | 3.75 | 29.62 | 2.1 | 9.26 | 57.48 | 5.26 | 6.57 |

| Dataset | Lang | #rows | Method | rouge-1 | | | rouge-2 | | | rouge-l | | | Final |
|---|---|---|---|---|---|---|---|---|---|---|---|---|---|
| | | | | f | p | r | f | p | r | f | p | r | Metric |
| internet | id | 18 | Peg_multi_news | 9.14 | 14.1 | 9.03 | 0.23 | 0.34 | 0.23 | 10.66 | 14.09 | 10.33 | 6.68 |
| archaeology | id | 46 | Peg_multi_news | 8.18 | 17.96 | 5.73 | 0.2 | 0.56 | 0.13 | 9.75 | 17.49 | 7.18 | 6.04 |

| Dataset | Lang | #rows | Method | rouge-1 | | | rouge-2 | | | rouge-l | | | Final |
|---|---|---|---|---|---|---|---|---|---|---|---|---|---|
| | | | | f | p | r | f | p | r | f | p | r | Metric |
| internet | id | 18 | Peg_wikihow | 10.88 | 36.95 | 7.72 | 3.03 | 14.89 | 1.79 | 12.54 | 38.2 | 8.33 | 8.82 |
| archaeology | id | 46 | Peg_wikihow | 7.37 | 41.53 | 4.19 | 2.14 | 14.07 | 1.22 | 9.19 | 43.11 | 5.29 | 6.23 |

| Dataset | Lang | #rows | Method | rouge-1 | | | rouge-2 | | | rouge-l | | | Final |
|---|---|---|---|---|---|---|---|---|---|---|---|---|---|
| | | | | f | p | r | f | p | r | f | p | r | Metric |
| internet | id | 18 | PN_cnndm | 15.99 | 36.68 | 12.8 | 11.74 | 27.37 | 9.32 | 18.75 | 41.54 | 14.36 | 15.49 |
| archaeology | id | 46 | PN_cnndm | 14.69 | 62.68 | 8.74 | 11.64 | 52.35 | 6.9 | 19.95 | 68.8 | 12.18 | 15.42 |

| Dataset | Lang | #rows | Method | rouge-1 | | | rouge-2 | | | rouge-l | | | Final |
|---|---|---|---|---|---|---|---|---|---|---|---|---|---|
| | | | | f | p | r | f | p | r | f | p | r | Metric |
| internet | id | 18 | PN_squad_qg | 2.14 | 15.54 | 1.2 | 0.56 | 2.62 | 0.32 | 3 | 15.83 | 1.73 | 1.9 |
| archaeology | id | 46 | PN_squad_qg | 2.36 | 26.83 | 1.26 | 1.04 | 9.41 | 0.56 | 3.27 | 26.51 | 1.78 | 2.22 |
| | | | MAX | 20.42 | 85.32 | 13.54 | 13.99 | 60.1 | 9.32 | 27.96 | 86.75 | 18.9 | 20.63 |

In general, from Table 3 and Table 4, for the top best pre-trained models, even until ROUGE-L-F, the precisions scores are high. The high scores of precisions indicate fewer repetitions in the reference summary, and there are many individual words/bigrams/LCSs instead. Table 3 shows that T5-Large achieved the best performance for the Wikipedia English dataset, by average considering the dataset, followed by ProphetNet-CNNDM and Pegasus-XSum. For the Wikipedia Indonesian dataset (Table 4), the T5-Large achieves the best performance, followed by Pegasus-XSum and ProphetNet-CNNDM. The rationale for these situations is even in an abstract summarization, the scores of their extractive coverage, density, and compression have a significant effect on their ROUGE performance. The T5-Large showed high fragmental coverage and high-density scores compared to other pre-training models (Pegasuses and ProphetNets). See Table 2(a) and Table 3; Table 2(b) and Table 4. A higher compression score in any summary means the summary captures many essential sights of the Wikipedia body text. All scores in Table 2(a) and Table 2(b) are normalized using the number of Wikipedia pages of the corresponding dataset.

**Table 5.** Examples of summary results (in English).

| Title | asymmetric digital subscriber line |
|---|---|
| T5-Large | ADSL differs from the less common symmetric digital subscriber line (SDSL) bandwidth and bit rate are said to be asymmetric, meaning greater toward the customer premises (downstream) than the reverse (upstream). in many places, it is the most common type offered to home users. |
| Pegasus-XSum | Asymmetrical digital subscriber line (ADSL) is a type of high-speed Internet service. |
| ProphetNet-CNNDM | ['in adsl, bandwidth and bit rate are said to be asymmetric, meaning greater toward the customer premises ( downstream ) than the reverse ( upstream ). [X_SEP] providers usually market adsl as an internet access service primarily for downloading content from the internet, but not for serving content accessed by others.'] |
| Pegasus-Multi_News | The first episode of Breaking Bad aired last night, and to celebrate, AMC is bringing the series to the big screen. "The only way to see Breaking Bad in IMAX is with our IMAX big-screen TVs," the network tweeted this morning. The Los Angeles Times notes that it's the first time a TV series will be shown in IMAX theaters. |
| Pegasus-Wikihow | Get to know what ADSL is and what it isn't.This article is written in the form of a guide for people who are new to the Internet. |
| ProphetNet-XGlue-NTG | ADSL: Asymmetric digital subscriber line |
| ProphetNet-Squad-QG | ['how does adsl differ from sdsl?'] |

Other additional factors that influence the ROUGE score are: 1) The goal of the pre-training. For example, the ProphetNet-Squad-QG goal is for question-generation, ProphetNet-XGlue-NTG more suitable for title summary purpose; 2) The dataset's characteristics. For example, the Pegasus-Multi_News gives too many biases in the output. Pegasus-Multi_News too memorizes its pre-trained article news dataset and not dub novel words properly. 3) The dataset used for testing the pre-trained model. The Wikipedia dataset we used in this paper describes the meaning of terminology. When we compare two methods, Pegasus-XSum and Pegasus-Wikihow, against the gold standard (viz the default Wikipedia-API), the summary results showed the ROUGE scores of the Pegasus-XSum by natural gives higher score than the one of the Pegasus-Wikihow. The rationale of this is the Pegasus-XSum is a pre-trained model using a wide variety of professionally written BBC articles. The Pegasus-Wikihow is the one pre-trained using an extensive collection instructions dataset (and not a collection describing the definition of a term). However, from evaluation (https://drive.google.com/drive/folders/152XMSCU3ctshB2BvzEQHY0vo6xkZ9gOl?usp=sharing, Table 5, Table 6), we can see the summaries produces by Pegasus-Wikihow also well capture important information of the whole reading. 4) The cross-lingual function. The function does not perform well in the ProphetNet in handling language other than English or Chinese. Even though the summary output when we tested the ProphetNets with the Wikipedia Indonesian language is well captured but sounds wording, they mix the two languages, some English words, and others in the Indonesian language. It needs postprocessing to translated the summary output into the Indonesian language before we compute its ROUGE scores. The rationale for this is because of the number of instances of Indonesian language used in the cross-lingual dataset not adequately enough. Theoretically, the two combined translated engines' performance strictly bound the pivoting framework results, especially to weaker ones. Train-infer-train procedure [7] can help implement zero-shot translation in the ProphetNet. We give ideas of future research implementation of the train-infer-train procedure on the zero-shot part at https://drive.google.com/drive/folders/1z09D2-4arE6aOQZxgxcwMVF41xMH4EEH?usp=sharing.

Table 6. Examples of summary results (in Indonesian language).

| Title | Akses Internet putar-nomor |
|---|---|
| T5-Large | Internet putar-nomor adalah teknologi informasi untuk akses Internet dengan menggunakan jaringan telepon tetap orang bergerak. |
| Pegasus-XSum | akses internet putar - nomor ( bahasa inggris : dial - up connection ) istilah teknologi informasi untuk akses internet menggundang jaringan telepon tetap atau telepon bergerak . |
| ProphetNet-CNNDM | [ ' adalah istilah teknologi informasi untuk akses internet . [ x _ sep ] dengan menggunakan jaringan telepon tetap atau telepon bergerak . ' ] |
| Pegasus-Wikihow | setagai internet putar - nomor ( bahasa inggris ) atau telepon tetap atau telepon bergerak , kamu . |
| Pegasus-Multi_News | Presiden Indonesia Susilo Bambang Yudhoyono telah memerintahkan peninjauan kembali aturan dan prosedur telekomunikasi negara itu setelah kegagalan jaringan besar-besaran yang menyebabkan ratusan ribu orang Indonesia tidak memiliki internet selama berhari-hari, … |
| ProphetNet-XGlue-NTG | cara mendapatkan tampilan baru di internet |
| ProphetNet-Squad_QG | [ ' what do teknologi informasi untuk ? ' ] |

**The Time Performance.** Fig. 1 shows the total CPU time against the Wall time for each Wikipedia dataset. From Fig. 1, we showed that T5-Large and Pegasus-XSum consume more time than other pre-trained model (ProphetNet). In general, the CPU time is linear to the number of Wikipedia pages processed in each dataset. It is showed that pre-trained models with high scores consume more CPU time than other slightly faster Transformers models.

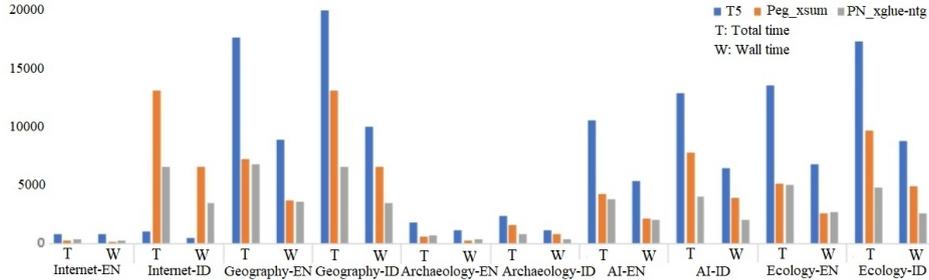

**Fig. 1.** A general overview of CPU time compares to the wall time. Time in seconds.

## 5     Conclusions and Prospects for Further Research

We surveyed several recent abstract summarization methods. From the survey, although T5-Large and Pegasus-XSum pre-trained models are top-2 excellent in summarization, another pre-trained model such as the ProphetNet (ProphetNet-CNNDM) also useful for summarization after postprocessing/translation, especially for any language other than English/Chinese.

The most significant factors that influence ROUGE performance are coverage, density, and compression. The higher the scores, the better the performance. Other factors that influence the ROUGE scores are the pre-training goal, the dataset's characteristics, the dataset used for the experiment (for testing the pre-training model), and the cross-lingual function. Several suggestions to improve this paper's limitation are: 1) assure that the dataset used for the pre-training model must sufficiently large, contains adequate instances for handling cross-lingual purpose; 2) Advanced process (finetuning) shall be reasonable. For cross-lingual summarization, the bilingual approach or cross-lingual strategy will be optional. We recommend using the large dataset consists of comprehensive coverage of topics from many languages before implementing advanced processes such as the train-infer-train procedure to the zero-shot translation in the training stage of the pre-training model.

## References

1. Cheng, J., Dong, L., Lapata, M.: Long short-term memory-networks for machine reading. arXiv preprint arXiv:1601.06733 (2016).
2. Dong, L., Yang, N., Wang, W., Wei, F., Liu, X., Wang, Y., Gao, J., Zhou, M., Hon, H.-W.: Unified language model pre-training for natural language understanding and generation. arXiv preprint arXiv:1905.03197 (2019).
3. Fabbri, A., Li, I., She, T., Li, S., Radev, D.: Multinews: A large-scale multi-document summarization dataset and abstractive hierarchical model. In: Proceedings of the 57th Annual Meeting of the ACL, pp. 1074–1084, Florence, Italy (2019).
4. Grusky, M., Naaman, M., Artzi, Y.: NEWSROOM: A Dataset of 1.3 Million Summaries with Diverse Extractive Strategies. In: Proc. Of NAACL-HLT 2018, pp. 708-719 (2018).

5. Hermann, K. M., Kocisky, T., Grefenstette, E., Espeholt, L., Kay, W., Suleymen, M., Blunsom, P.: Teaching machines to read and comprehend. In: Advances in neural information processing systems, pp. 1693-1701 (2015).
6. Koupaee, M., Wang, W. Y.: Wikihow: A large scale text summarization dataset. arXiv preprint arXiv:1810.09305 (2018).
7. Lakew, S. M., Lotito, Q. F., Negri, M., Turchi, M., Federico, M.: Improving Zero-Shot Translation of Low-Resource Languages. arXiv preprint arXiv:1811.01389v1 (2018).
8. Lewis, M., Liu, Y., Goyal, N., Ghazvininejad, M., Mohamed, A., Levy, O., Stoyanov, V., Zettlemoyer, L.: Bart: Denoising sequence-to-sequence pre-training for natural language generation, translation, and comprehension. arXiv preprint arXiv: 1910.13461 (2019).
9. Liang, Y., Duan, N., Gong, Y., Wu, N., Guo, F., Qi, W., Gong, M., Shou, L., Jiang, D., Cao, G., Fan, X., Zhang, R., Agrawal, R., Cui, E., Wei, S., Bharti, T., Qiao, Y., Chen, J.-H., Wu, W., Liu, S., Yang, F., Campos, D., Majumder, R., Zhou, M.: XGLUE: A New Benchmark Dataset for Cross-lingual Pre-training, Understanding and Generation. arXiv preprint arXiv: 2004.01401 (2020).
10. Lin, C.-Y.: ROUGE: A Package for Automatic Evaluation of Summaries. In: Text summarization branches out, pp. 74-81. ACL, Barcelona, Spain (2004).
11. Narayan, S., Cohen, S. B., Lapata, M. Don't give me the details, just the summary! Topic-aware convolutional neural networks for extreme summarization. In: Proceedings of the 2018 Conference on EMNLP, pp. 1797-1807. ACL, Brussels, Belgium (2018).
12. Qi, W., Yan, Y., Gong, Y., Liu, D., Duan, N., Chen, J., Zhang, R., Zhou, M.: ProphetNet: Predicting Future N-gram for Sequence-to-Sequence Pre-training. arXiv preprint arXiv:2001.04063v3 (2020).
13. Raffel, C., Shazeer, N., Roberts, A., Lee, K., Narang, S., Matena, M., Zhou, Y., Li, W., Liu, P.J.: Exploring the Limits of Transfer Learning with a Unified Text-to-Text Transformer. Journal of Machine Learning Research 21, 1-67 (2019).
14. Rajpurkar, P., Zhang, J., Lopyrev, K., Liang, P.: SQuAD: 100,000+ Questions for Machine Comprehension of Text. arXiv preprint arXiv:1606.05250 (2016).
15. Rothe, S., Narayan, S., Severyn, A.: Leveraging pre-trained checkpoints for sequence generation tasks. arXiv preprint arXiv: 1907.12461 (2019).
16. Song, K., Tan, X., Qin, T., Lu, J., Liu, T.-Y.: MASS: Masked sequence to sequence pre-training for language generation. arXiv preprint arXiv:1905.02450 (2019).
17. Vaswani, A., Shazeer, N., Parmar, N., Uszkoreit, J., Jones, L., Gomez, A. N., Kaiser, L., Polosukhin, I.: Attention is all you need. In: Proceedings of the 31st International Conference on NIPS, pp. 6000–6010. NIPS 2017, Long Beach, CA (2017).
18. Wang, A., Singh, A., Michael, J., Hill, F., Levy, O., Bowman, S.: GLUE: A Multi-Task Benchmark and Analysis Platform for Natural Language Understanding. ICLR (2018). In: Proceedings of the 2018 EMNLP, pp. 353-355. ACL, Brussels, Belgium (2018).
19. Zhang, J., Zhao, Y., Saleh, M., Liu, P. J.: PEGASUS: Pre-training with Extracted Gap-sentences for Abstractive Summarization. In: Proceedings of the 37th International Conference on Machine Learning, pp. 11328-11339. PMLR 119, Vienna, Austria (2020).